\documentclass[final,5p,times,twocolumn]{elsarticle}

\usepackage{amssymb}

\usepackage{booktabs}
\usepackage{algorithm}
\usepackage{algorithmic}
\usepackage{siunitx}

\usepackage{hyperref}
\usepackage{multirow}
\usepackage{multicol} 
\usepackage{subfigure}
\usepackage{graphicx} 
\usepackage{epstopdf}
\usepackage{dsfont}
\usepackage{amsmath}

\usepackage{bbding} 

\usepackage[T1]{fontenc}

\usepackage{threeparttable}

\usepackage{color}

\usepackage{color}
\usepackage{xcolor}
\definecolor{wrong}{rgb}{.8,.349,.1}
\definecolor{right}{rgb}{.3,.7,.1}
\usepackage{newfloat}
\usepackage{listings}
\usepackage{amsmath}
\usepackage{amssymb}

\makeatletter
\newcommand\figcaption{\def\@captype{figure}\caption}
\newcommand\tabcaption{\def\@captype{table}\caption}
\makeatother

\journal{Elsevier}

\begin{document}

\begin{frontmatter}



\title{ Cross-head mutual Mean-Teaching for semi-supervised medical image segmentation}



\author[a]{Wei Li}
\ead{leesoon@bupt.edu.cn}
\author[a]{Ruifeng Bian}
\ead{ruifeng_bian@bupt.edu.cn}
\author[a]{Wenyi Zhao}
\ead{zwy@bupt.edu.cn}
\author[a]{Weijin Xu}
\ead{xwj1994@bupt.edu.cn}

\author[a]{Huihua Yang \corref{cor1}}
\ead{yhh@bupt.edu.cn}

\address[a]{School of Artificial Intelligence, Beijing University of Posts and Telecommunications, Beijing, 100876, China }

\cortext[cor1]{Corresponding author: Huihua Yang}

\begin{abstract}

Semi-supervised medical image segmentation (SSMIS) has witnessed substantial advancements by leveraging limited labeled data and abundant unlabeled data. Nevertheless, existing state-of-the-art methods  encounter challenges in accurately predicting labels for the unlabeled data, giving rise to disruptive noise during training and susceptibility to erroneous information overfitting.  Moreover, applying perturbations to inaccurate predictions further reduces consistent learning. To address these concerns, we propose a novel \textbf{C}ross-head \textbf{m}utual \textbf{m}ean-\textbf{t}eaching Network (CMMT-Net) incorporated strong-weak data augmentation, thereby benefitting both self-training and consistency learning.  Specifically, our CMMT-Net  consists of both  teacher-student peer networks with a share encoder and  dual slightly different decoders, and the pseudo labels generated by one mean teacher are adopted to  supervise the other student to achieve a mutual consistency between two branches. Furthermore, we propose  mutual virtual adversarial training (MVAT) to smooth the decision boundary and enhance feature representations. enhance feature representations.  To diversify the consistency training samples, we employ Cross-Set CutMix strategy, which also helps address distribution mismatch issues. Notably,  CMMT-Net simultaneously implements  data, feature, and network perturbations,  amplifying   model diversity and generalization performance. Experimental results on three publicly available datasets  indicate that our approach yields remarkable improvements over previous  state-of-the-art (SOTA) methods across  various semi-supervised scenarios. Code and logs will be available at \url{https://github.com/Leesoon1984/CMMT-Net}.

\end{abstract}



\end{frontmatter}


\section{Introduction}

Automated and robust medical image segmentation plays a critical role in facilitating computer-aided diagnosis (CAD), providing clinicians with a solid foundation for illness diagnosis by accurately segmenting regions such as cells, tissues, and organs. Recent advancements in deep learning have enabled Convolutional Neural Networks (CNNs) to achieve remarkable proficiency in a wide range of medical image segmentation tasks. However, clinical applications face a recurring challenge due to the limited availability of training data. On the one hand, annotating medical images at the pixel or voxel level  imposes heightened demands compared to the annotation of natural images, necessitating meticulous time investments and the participation of medical experts equipped with clinical experience. On the other hand, clinical scenarios typically involve abundant unlabeled data in contrast to the scarcity of labeled data.  Consequently, leveraging labeled and unlabeled medical data through techniques like semi-supervised methods  to enhance accuracy has gained significant research attention.

The current state-of-the-art (SOTA) semi-supervised learning methods can be broadly categorized into two main categories: pseudo training (self-training)~\cite{chen2021semi} and consistency regularization training~\cite{yu2019uncertainty,li2020shape, wu2022mutual, zhang2017deep}.  In pseudo-label training, pseudo-labels are incorporated into unlabeled images obtained from the segmentation model, guiding its learning process. Subsequently, the segmentation model undergoes retraining using both labeled and unlabeled data. However, conventional self-training methods suffer from inherent limitations, such as the phenomenon of confirmation bias~\cite{arazo2020pseudo}, where the pseudo-label noise can accumulate and significantly impact the entire training process.  As an extension of self-training, co-training~\cite{wu2022mutual} introduces the notion of multiple individual learners benefiting from each other, the key challenge here is to devise a strategy that mitigates the risk of different sub-networks collapsing into each other. 

On the other hand, consistency regularization training techniques exploit unlabeled data during training by enforcing prediction consistency across various perspectives, including data-level, feature-level, and network-level views. By applying perturbations to the input image using data augmentation ~\cite{DBLP:conf/iclr/ZouZZLBHP21} or to the feature space via noise injection~\cite{Ouali_2020_CVPR}, this approach prompts the network to generate consistent predictions for given unlabeled images subjected to various augmentations.  Moreover, Network perturbation offers an alternative avenue for acquiring diverse perspectives,  facilitating alignment in predictions across multiple models that are initialized differently. A typical network perturbation method,  CPS~\cite{chen2021semi},  involves feeding the same image into two independently initialized networks and leverages the pseudo labels generated from one branch to supervise the other. However, it's important to note that consistency learning has a notable vulnerability: it assumes accurate predictions for unlabeled images, presuming alignment with the true classification decision boundary.  In practical scenarios, even SOTA methods frequently struggle to fulfill this assumption, which can lead to a potentially flawed training signal for consistency learning. This challenge is particularly pronounced in network perturbation-based consistency learning, where erroneous predictions from one model can detrimentally impact the training of its counterpart, and vice versa.

 To fully take the merit of consistency regularization and co-training, some works~\cite{wu2022mutual, huang2022complementary, wang2023mcf, huang2022complementary, wang2023mcf, xu2023ambiguity, bai2023bidirectional}  adopt various perturbations to achieve consistency learning. MC-Net~\cite{wu2021semi} employs two distinct segmentation heads to facilitate  consistency learning via mutually supervising each other through the  sharpened prediction probabilities. To diversify the feature-level perturbations, MC-Net+~\cite{wu2022mutual} incorporates multiples decoders to regularize the model training. In addition, MCF~[28] introduces an innovative mutual correction framework (MCF) aimed at investigating network bias correction. AC-MT~\cite{xu2023ambiguity}  comprehensively introduces a family of plug-and-play strategies designed for selecting ambiguous targets. To address challenging regions more effectively, CC-Net~[21] departs from the shared encoder design and instead employs two complementary auxiliary networks to introduce inter-model perturbations, subsequently enforcing consistency among these three models. However, it's important to note that this strategy may require an increased number of model parameters, as it involves the simultaneous training of three distinct models. Different from these methods, BCP~\cite{bai2023bidirectional} adopts Bidirectional Copy-Paste between labeled and unlabeled data to enhance the model generalization by increasing diversity of unlabeled data. However, the challenge of incorporating data-level, feature-level, and network-level perturbations into a cohesive framework to achieve multi-view consistency learning has not been comprehensively investigated. For instance, as noted in \cite{sss21simple,zhao2022augmentation}, excessive data perturbations can adversely affect the distribution of the original data and, consequently, lead to performance degradation. Furthermore, medical image data exhibits intricate and diverse characteristics, which pose challenges for SSMIS methods, thereby potentially hindering their ability to achieve robust generalization across various datasets.

 To address these issues, in this paper, a novel Cross-head mutual mean-teaching Network (CMMT-Net) incorporated with weak and strong augmentation is proposed, as is illustrated in Fig.~\ref{fig:framework}. Our CMMT-Net is built upon the cross-head co-training paradigm~\cite{wu2021semi}, which  consists of a shared encoder and slightly different dual decoders. Additionally, we incorporate an  auxiliary two branch  mean teacher model to  capture information  and  offer supplementary supervision. During the training process, weakly augmented images are fed to the teacher to increase confidence in the predictions, which are then utilized to supervise the strongly augmented  predictions from the other  student.  This approach allows the student to be trained on more diverse and challenging samples, and  the pseudo labels generated by the teacher will be more accurate and robust, allowing for the  utilization of more challenging perturbations that combine input image, feature, and network perturbations to enhance consistency learning. Specifically, our shared encoder introduces constraints on diverse learners, thereby preventing them from converging in opposing directions. In addition, to prevent co-training from degrading into self-training, we employ Cross-Set CutMix to achieve data-level perturbations. This augmentation technique involves CutMix~\cite{yun2019cutmix} on both the labeled and unlabeled data, thereby facilitating cross-view consistency learning while narrowing the domain gaps between them~\cite{wang2019semi}.
 We also propose  mutual virtual adversarial training (MVAM) on both the labeled and unlabeled data to smooth the decision boundary. Note that the generation of adversarial noises relies on the output distribution between the teacher-student decoders, which incorporates  complementary information and yields adversarial noise with less confirmation bias~\cite{arazo2020pseudo}. Extensive experiments verify that the proposed CMMT-Net framework benefits from the designed perturbations. In summary, our contributions are as follows:

\begin{itemize}
	\item This paper proposed a cooperative Cross-head mutual mean-teaching Networks (CMMT-Net) to synergistically integrate data-level, feature-level, and network-level perturbations for enhancing SSMIS tasks.
	\item We introduce a  new type of data perturbation strategy called MVAT, which utilizes the adversarial noise learned from the teacher to augment the inputs  of the student model, thereby benefitting the model in decision boundary smoothing and better
    representation learning. 
	\item A Cross-Set CutMix strategy is proposed to diversify the training samples and reduce the domain gaps between the labeled and unlabeled data.
	\item Extensive experiments on three public data demonstrate that the proposed CMMT-Net surpasses existing methods by a significant margin  across all labeled data ratios, establishing a new SOTA in the field of SSMIS.
\end{itemize}

\section{Related Works}

\subsection{Semi-supervised Learning}

Semi-supervised learning (SSL) has been extensively employed in numerous computer vision tasks by  harnessing unlabeled and labeled images together for training. Existing SSL approaches can be divided into the following categories: Pseudo-labeling~\cite{lee2013pseudo} and consistency regularization training~\cite{sajjadi2016regularization,tarvainen2017mean}. Pseudo-labeling-based  methods leverage pseudo labels generated by the model's predictions to train on unlabeled data. utilize pseudo labels generated by the model predictions to train the models on unlabeled data. On the other hand, consistency regularization methods introduce perturbations at different levels, i.e., data level, feature level, and network level,  to enforce  consistency among the predictions of various views.  

{\textbf{Pseudo-labeling: } Pseudo-labeling-based methods typically involve assigning pseudo-labels to unlabeled data using either a fixed or dynamic threshold. These pseudo-labeled data are subsequently integrated with manually annotated data for further training and refinement of the model. For example, ACPL~\cite{liu2022acpl} enhances the accuracy of pseudo-labels through ensembling classifiers, and adopts an anti curriculum training approach. Noise Student~\cite{Xie2019SelfTrainingWN}  employs an iterative learning approach, where pseudo-labels are generated using an updated teacher network, and the student network is directed to learn from the entire dataset. In addition, various approaches \cite{sohn2020fixmatch, cascante2021curriculum, zhang2021flexmatch, kim2021distribution} provide sample selection strategies to generate pseudo-labels.  FixMatch \cite{sohn2020fixmatch}  utilizes a predefined threshold to select pseudo-labeled samples with highly confident predictions  for the purpose of consistency learning. Instead of relying on a fixed threshold, \cite{zhang2022boostmis, zhang2021flexmatch, wang2022freematch, chen2023softmatch} employ a dynamic threshold to select pseudo-labeled samples, allowing for a trade-off between the quantity and quality of the pseudo-labels. Unlike these methods, we face challenges in relying on a probability threshold as a reliable criterion for selecting clean pseudo-labeled data. This difficulty arises because unlabeled data, whether with correct or incorrect pseudo-labels, often exhibit similar probability distributions. Instead, we adopt the pseudo-labels generated by the weakly augmented predictions from the  teacher to supervise the strongly augmented predictions from the student.

    {\textbf{Consistency-based methods:} The primary objective of consistency regularization is to reduce output discrepancies when different perturbations are applied to various views of unlabeled data. It is imperative to establish high-quality consistency targets during training to attain optimal performance.  A representative  method, Mean-Teacher (MT)~\cite{tarvainen2017mean} enforces similarity between predictions of the student model and its momentum teacher model.  However, MixMatch~\cite{berthelot2019mixmatch} proposes a technique to reduce the discrepancy among multiple samples that are augmented using Mixup. To harness the full potential of unlabeled data, SimMatch~\cite{zheng2022simmatch} resorts to contrastive learning by encouraging augmented views of the same instance to exhibit consistent class predictions and similar relationships with regard to other instances. Additionally, \cite{wang2023deep} introduces deep semi-supervised multiple instance learning with self-correction. However,  these  methods  ignore  the  interactions  between subnets and also cannot correct the biases of the network itself. Unlike  the aforementioned methods, we employ a dual-network approach to actively enforce consistency under data-level, feature-level, and network-level perturbations. This substantially expands the representation space and enhances the generalization of the model.

 \begin{figure*}[!ht]
	\centering 
	\includegraphics[scale=0.4]{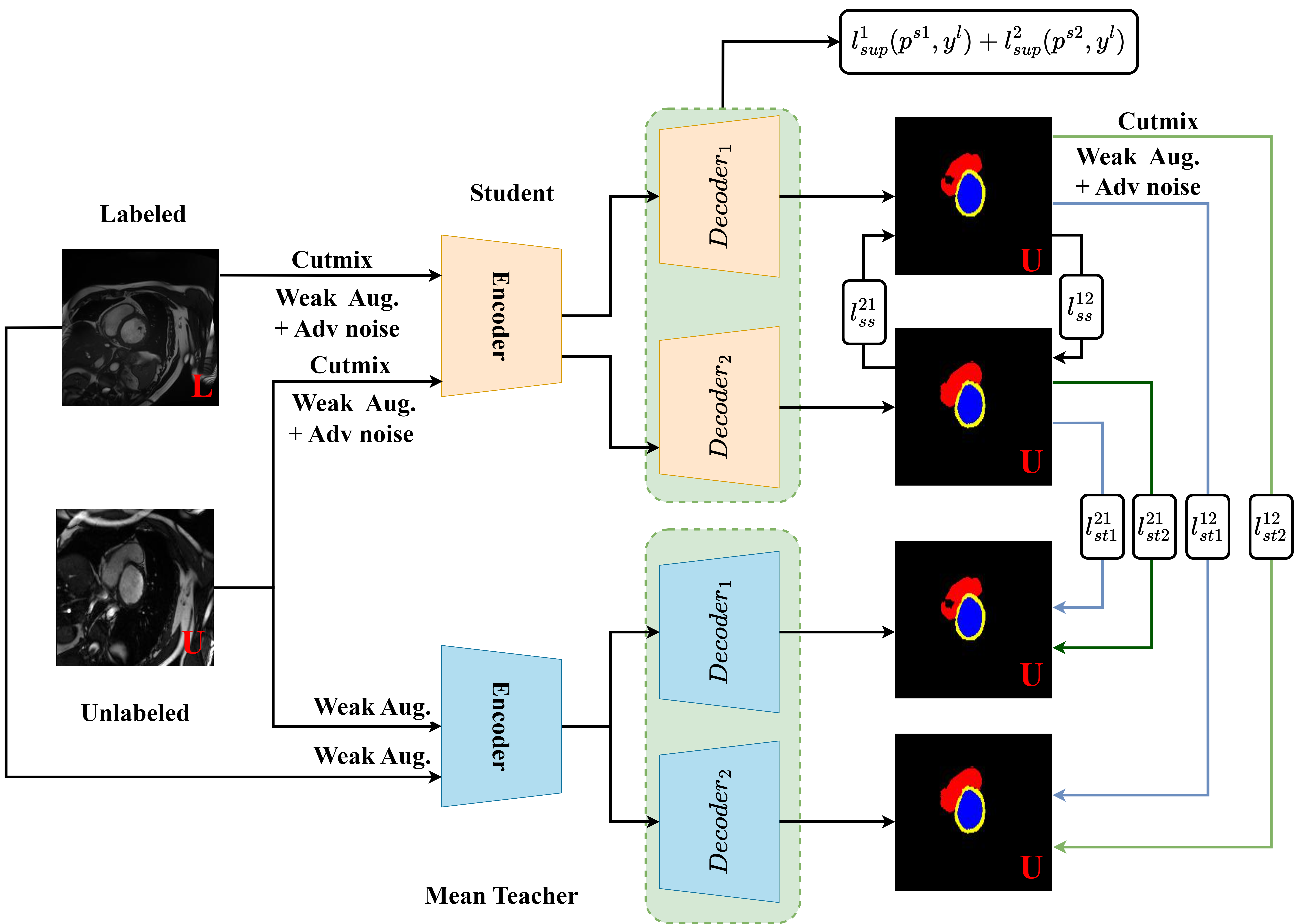}
	\caption{ An overview of the proposed Cross-head Mutual mean-teaching network (CMMT-Net) architecture. Our CMMT-Net consists of teacher-student peer networks,  each featuring  a shared encoder and dual different decoders  that facilitate interactions between the decoders. More specifically, to train the student model, we use first head as an example.  Weakly augmented labeled and unlabeled images are fed to the shared encoder and corresponding $Decoder_1$ to generate predictions, then they will be supervised by Ground Truth (GT) with the loss $l^{1}_{sup}$ and the predictions of the second $Decoder_2$  with the loss $l^{12}_{ss}$, respectively. Furthermore,  weakly augmented are fed to the shared teacher encoder and corresponding $Decoder_2$ to generate more accurate predictions, then the predictions are mixed to supervise the CutMixed image predictions from the other student $Decoder_1$ with the loss $l^{12}_{st2}$. In addition, inputs perturbed by adding adversarial noise (adv noise) estimated by the second teacher are fed to the student $Decoder_1$ to achieve consistency learning with the loss $l^{12}_{st1}$. }
	\label{fig:framework}
\end{figure*}

\subsection{Semi-Supervised Medical Image Segmentation}

To alleviate the expenses associated with pixel/volume-level annotation while preserving   high accuracy performance, recent developments have introduced various semi-supervised methods for medical image segmentation tasks.  Among these methods, the Mean-Teacher approach~\cite{tarvainen2017mean} and its extensions~\cite{yu2019uncertainty,li2020shape} have received considerable attention. UA-MT~\cite{yu2019uncertainty} utilizes a mean-teacher framework to steer the student model towards producing predictions consistent with the teacher model through an uncertainty-aware training approach.  Meanwhile, SASSNet~\cite{li2020shape} proposes a geometric shape constraint on the segmentation outputs to improve accuracy. Introducing an extra task-level constraint, DTC~\cite{luo2021semi} adopts a dual task-consistency framework. Furthermore, the introduction of Uncertainty Rectified Pyramid Consistency~\cite{luo2022semi} aims to diminish disparities  between pyramid predictions and their aggregate average. MC-Net+\cite{wu2022mutual} decreases the disparity among the outputs of multiple decoders through mutual consistency. To further explore the difficult regions, CC-Net~\cite{huang2022complementary} departs from the shared encoder design and incorporates two additional complementary auxiliary networks to achieve inter-model perturbations and enforce consistency among these three models. CTCL~\cite{li2022collaborative} introduces an extra transformer branch and encourages prediction consistency between the CNN model and the Transformer model to enable the model to benefit from the two learning paradigms. BCP~\cite{bai2023bidirectional} joint processes the labeled and unlabeled data using a copy-paste strategy and encourages the prediction consistency between the teacher model and the student model. AC-MT~\cite{xu2023ambiguity} presents ambiguity-selective consistency regularization for mean-teacher SSMIS.  Recently, contrastive learning methods are widely to apply for SSMIS  tasks. Semi-CM~\cite{zhang2023multi}  presents a semi-supervised contrastive mutual learning segmentation framework via incorporating a novel area-similarity contrastive (ASC) loss that leverages cross-modal information and ensures prediction consistency between different modalities. Wang et al. \cite{wang2021exploring} extends image-level contrastive learning to the pixel level to enhance segmentation performance. SS-Net~\cite{wu2022exploring} utilizes a prototype-based strategy to disentangle the feature of different classes, thereby promoting class-level separation and leveraging low-entropy regularization for model training. Unlike classification tasks that employ a contrastive loss to learn global-level representations from unlabeled images, Chaitanya et al. \cite{chaitanya2023local} adapt local contrastive loss to learn pixel-level features that are valuable for segmentation. In addition, Wang et al. \cite{wang2021deep} introduces Virtual Adversarial Training (VAT) \cite{miyato2018virtual} with consistency regularization for semi-supervised medical image classification. Compared to VAT, a method based on adversarial training, our proposed Mutual Virtual Adversarial Training (MVAT) generates adversarial noises  based on the output distribution between two different perturbed views, which incorporates complementary information and produces less confirmation-biased~\cite{arazo2020pseudo} adversarial noises.

\section{Methodology} 

\subsection{Problem Definition}

The objective of the SSMIS task is to enhance segmentation performance on the unseen images by leveraging both limited labeled data and amounts of unlabeled data.  Let $\mathcal{D}^{l}=\{(x_i,y_i)|i=1,...,N\}$ represent a small labeled dataset with $N$ annotated images, where $x_i\in\mathcal{X}_l$ and $y_i\in\mathcal{Y}_l$. Additionally, let $\mathcal{D}^{u}=\{x_j|j=1,...,M\}$ denotes a significantly larger unlabeled dataset with $M$ raw images ($M \gg N$), where $x_j \in \mathcal{X}_u$. 

To address the mentioned challenge, we propose the Cross-head Mutual Mean-Teaching Network (CMMT-Net), which comprises both teacher-student networks. In this architecture, the teacher and student models share the same network represented as $f$ with parameters $\theta$. The model is decomposed into a shared encode $g$ and $m (m \in \{ 1,2\})$ segmentation decoders $\{ h^{m} \}$ with different architectures, and can be written as $f^m=g \circ \{h^{m} \}$ with the parameter $\theta^m = \{ \theta_g, \theta_h^{m}\}$. Here the pseudo label is generated by $p^{m}(y|x) = h^m(g(x))$, then the pseudo label will be utilized to supervise the training of the other head. The mean teacher-student model is represented with the respective parameter superscripts: $\theta^{s1} = \{\theta_g^{s}, \theta_h^{s1}\}$ and $\theta^{s2} = \{\theta_g^{s}, \theta_h^{s2}\}$ for the dual-branch student, $\theta^{t1} = \{\theta_g^{t}, \theta_h^{t1}\}$ and $\theta^{t2} = \{\theta_g^{t}, \theta_h^{t2}\}$ for the dual-branch teacher.

\subsection{Overview of the Framework}

Fig.~\ref{fig:framework} provides an overview of the workflow for our proposed Cross-head Mutual Mean-Teaching Network (CMMT-Net) architecture. CMMT-Net is composed of teacher-student peer networks, where the student networks comprise a shared encoder and two dual decoders, and the teacher network is constructed using exponential moving average (EMA) of the parameters from the student network. During the training process, weakly augmented images are fed to the teacher to boost confidence in the predictions, and these predictions are then used to supervise the strongly augmented predictions from the other student. To enhance the diversity of the model and improve its generalization, our CMMT-Net incorporates perturbations at three different levels: data, feature, and network.  Details for each part are described as follows.

\noindent \textbf{Baseline student network.} 
Contrary to the vanilla U-Net~\cite{ronneberger2015u} or  V-Net~\cite{milletari2016v} which  feature  a single segmentation head, our student network $f^{sm}$ is build upon the MC-Net~\cite{wu2021semi, wu2022mutual}, which  is  composed   of  a  shared encoder $g^s$ and two slightly different decoders $h^{sm} (m \in \{ 1,2\})$. These decoders  differ in the up-sampling strategies: one employs the transposed convolution layer, while the other integrates the linear interpolation layer. This intentional distinction in decoding strategies introduces feature-level perturbations, contributing to the network's diversity.  In addition, compared with using separate individual models like CPS~\cite{chen2021semi} and CC-Net~\cite{huang2022complementary},  our approach allows for co-training with minimal additional parameters using the shared encoder. The shared encoder imposes constraints on different decoders, preventing them from converging in opposite directions. Consequently, more compact features can be learned, enhancing the  generalization of the model.

\noindent \textbf{Auxiliary mean teacher network.}
In addition to feature-level perturbations, we also delve into network-level diversity. To achieve this, we employ a straightforward approach based on the mean teacher strategy~\cite{tarvainen2017mean} to create  auxiliary mean-teacher networks $f^{tm}=g^{t} \circ h^{tm}$ $(m \in \{ 1,2\})$. The effectiveness of mean teacher networks in capturing historical information to enhance model performance has been demonstrated in prior studies~\cite{he2020momentum, grill2020bootstrap, tarvainen2017mean}. This approach does not require explicit optimization and introduces minimal computational overhead. The network structure of the teacher is the same as the that of the student, and the parameters of the teacher are updated by the corresponding student model using the exponential moving average (EMA), where the $\gamma$ controls the speed of the updates and $m \in[1,2]$ represents the index of the segmentation heads.

\begin{align}
    \theta^{tm} = \gamma \theta^{tm} + (1-\gamma) \theta^{sm}
    \label{eq: update teacher}
\end{align}

\noindent \textbf{Data-level Perturbation.}
Diversity is essential for the generalization ability of the models, the co-training may fall into the self-training when the diversity is limited. To further enlarge the perturbation spaces of the models, we explore to increase the diversity on data-level.
While data augmentation has been widely successful in semi-supervised tasks with natural images, its effectiveness in the context of medical images, especially in 3D, is an area where further exploration is needed. Many common data augmentation technologies, like ColorJitter~\cite{sohn2020fixmatch}, cannot be used directly on the medical images. We expect to find a common approach for the image augmentation at pixel-level. Inspired by the virtual adversarial training (VAT)~\cite{miyato2018virtual}, adding adversarial perturbation is beneficial for smoothing decision boundaries and tackling edge-distributed samples. To effectively harness both labeled and unlabeled data for the segmentation task, mutual virtual adversarial training (\textbf{MVAT}) is utilized to  estimate  the  adversarial  noise  using  the  more  accurate teachers  and then apply this estimated noise to the inputs of the student model, thereby benefitting the model in decision boundary smoothing and better representation learning. In addition, the generation of adversarial noises is based on the output distribution between two different networks, which incorporates complementary information and produces less confirmation-biased adversarial noises. The detailed design of the \textbf{MVAT} can be seen in the Section~\ref{VAMMT}. On top of the  virtual adversarial  data augmentations, we also extend the CutMix~\cite{yun2019cutmix} strategy for the 3D segmentation task and proposed mutual mean-teaching with CutMix. Note that this augmentation involves CutMix~\cite{yun2019cutmix} both on the labeled and unlabeled data, thereby facilitating cross-view consistency learning while narrowing the domain gaps between them~\cite{wang2019semi}, ultimately improving model generalization, more details can be seen in Section~\ref{CMMT-Mix}.

\subsection{Model Architecture}\label{model_labeled}

The proposed CMMT-Net consists of a shared encoder model $g$ and $m$ segmentation decoders $\{h^m\}$, which can learn various characteristics to be beneficial to each other. Given an input image $x_i \in D^{l}\cup  D^{u}$, the proposed framework produces two predictions:

\begin{equation}
	p_{i}^{1}=h^1\left(g(x_{i})\right); p_{i}^{2}=h^2\left(g(x_{i})\right)
\end{equation}
where $p_i^{m}$ denotes the prediction  of m-th segmentation head.

\textbf{ Supervised Learning.} First, we train the student model with the labeled data. Dice loss $\mathcal{L}_{\text {dice}}$ is used as follows:

\begin{equation}
	\mathcal{L}_{\text {sup }} = \sum_{x_i,y_i\in D^l}\mathcal{L}_{dice}\left(p_{i}^{s1}, y_{i}\right) + \mathcal{L}_{dice}\left(p_{i}^{s2}, y_{i}\right)
 \label{loss:sup}
\end{equation}
where $p_{i}^{s1}$ and $p_{i}^{s2}$ denotes the predictions of student model.

\textbf{Cross-head co-training.} The predictions between the  two segmentation heads have different properties, essentially in the output level, we follow the ~\cite{chen2021semi} to achieve Cross-head co-training (\textbf{CCT}) with hard pseudo labels. Based on the predictions of two segmentation heads, we employ argmax function to yield the hard pseudo labels as follows:

\begin{equation}
	\hat{y}_{i}^{sm}=\operatorname{argmax}\left(p_{i}^{sm}\right), m \in \{ 1,2\}
\end{equation}

Then, we jointly train the student network on the unlabeled data as follows:

\begin{equation}
	\mathcal{L}_{\text {ss}} =\sum_{x_i \in D^u}\mathcal{L}_{dice}\left(p_{i}^{s1}, \hat{y}_{i}^{s2}\right)+\mathcal{L}_{\text {dice }}\left(p_{i}^{s2}, \hat{y}_{i}^{s1}\right)
    \label{loss:cct}
\end{equation}

\subsection{Mutual Virtual adversarial Training} \label{VAMMT}

 \begin{figure}[!ht]
	\centering 
	\includegraphics[scale=0.14]{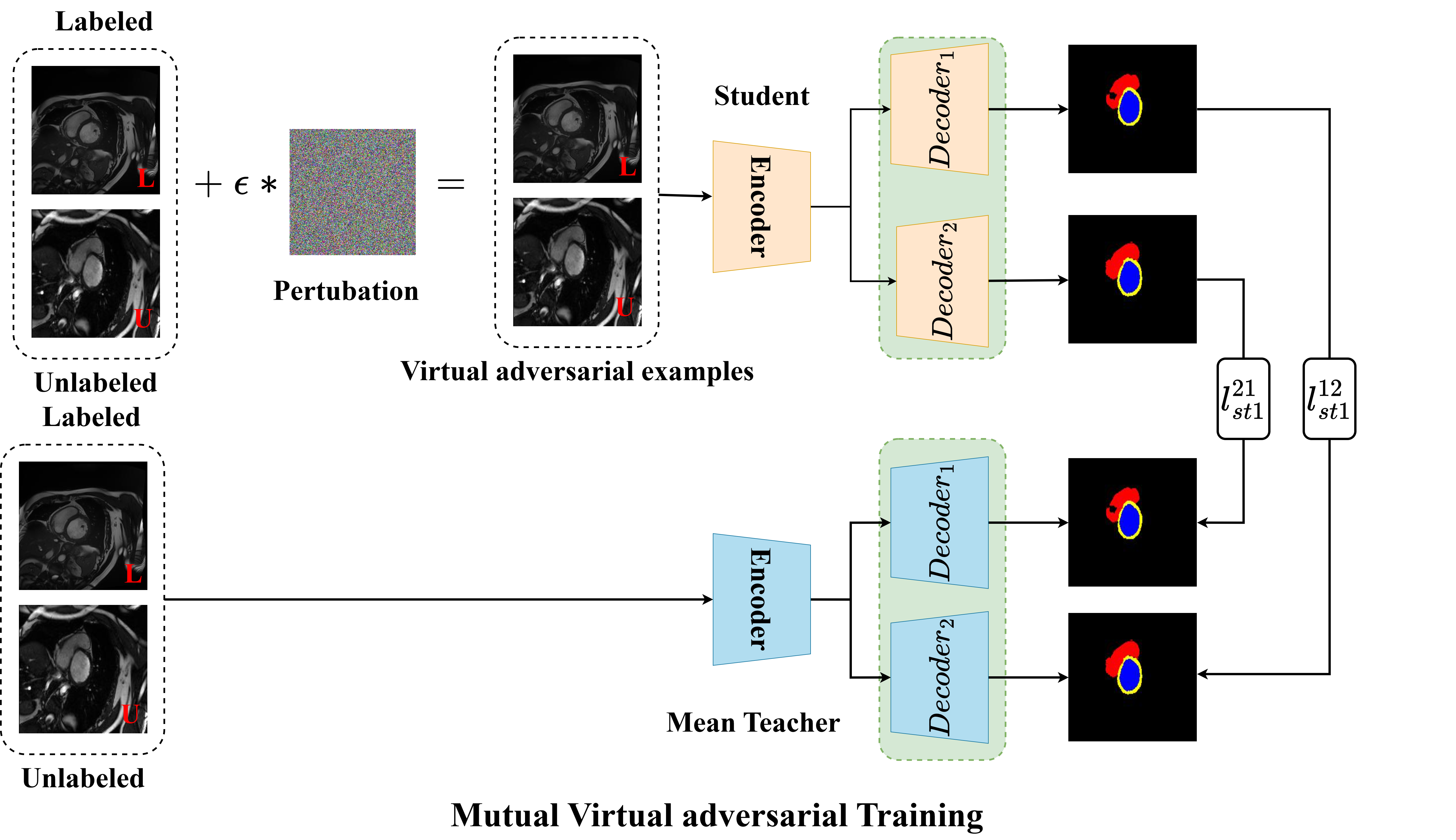}
	\caption{ The  framework of Mutual Virtual Adversarial Training (MVAT).}
	\label{fig:vammt}
\end{figure}

\textbf{Virtual adversarial training.}  Adversarial training ~\cite{Goodfellow2014GenerativeAN}  is designed to train the model to provide each input data a similar prediction to its neighbors in the adversarial direction, which makes the model robust against adversarial perturbation, ultimately improving model generalization capability. Different from the adversarial direction, virtual adversarial Training (VAT)~\cite{miyato2018virtual} can be defined on unlabeled examples, which seeks the direction that can most significantly alter the current inferred output distribution. VAT~\cite{miyato2018virtual}  has proven successful in numerous semi-supervised learning tasks. It leverages the concept of local distributional smoothness (LDS) to quantify the model's distributional robustness against virtual adversarial perturbations. Virtual adversarial examples are integrated by minimizing the following objective function:

\begin{equation}
	LDS(x; \theta)= \mathcal{D}[ q(y|x), p(y|x+r_{vadv}, \theta) ]
\end{equation}
where:

\begin{equation}
	r_{adv} := \mathop{\arg\max}_{r;\|r\| \leq \delta}\mathcal{D}[ q(y|x_{\ast}), p(y|x_{\ast}+r_{vadv}, \hat{\theta}) ]
 \label{vat}
\end{equation}
Here $\mathcal{D}(q,p)$  is used to compute the discrepancy between $q$ and $p$,  $x_{\ast}$ denotes either $x_l$ or $x_u$,  $r_{adv}$ is the adversarial perturbation, $\theta$ is a set of trainable parameters of the model, $\hat{\theta}$ is the current estimate of $\theta$, $\delta$ is a hyperparameter controlling the perturbation size. $p(y|x;\theta)$ is the output distribution of a model; $q(y|x)$ is the true distribution of the output label, which is unknown; $x+r_{adv}$ is the virtual adversarial example corresponding to the input sample $x_{\ast}$.

\textbf{Mutual Virtual adversarial Training.} However, we  contend that the local perturbations may be limited to further enhanced the generalization ability of the models. Besides, a common drawback of VAT is that it enforces consistency without  assessing the quality of these outputs.
Current  methods  estimate the adversarial noise using the same single network where  the  consistency  loss  will  be  applied~\cite{miyato2018virtual}, while the noisy included in the predictions by the student network can lead to confirmation bias~\cite{arazo2020pseudo}. Given that the student model tends to have less accurate predictions than the teacher model, this approach may not be conducive to effective training.  Therefore,  we propose  to  estimate  the  adversarial  noise  using  the  more  accurate teachers, and then apply this estimated noise to the input of the student model. To mitigate the impact of samples near the decision boundary, we introduce the Mutual Virtual Adversarial Training (\textbf{MVAT}) framework to expand the perturbation space and enhance the generalization ability of the models.

As depicted in Figure~\ref{fig:vammt}, we employ the predictions from the weakly augmented samples generated by the mean-teacher to provide cross-head supervision for the training of corresponding virtual adversarial examples. Specifically, we utilize the output $q^{t2}(y|x)$ from the teacher head $h^{t2}$ to generate adversarial noise for the first decoder $h^{s1}$, and the output $q^{t1}(y|x)$ from the teacher head $h^{t1}$ to generate adversarial noise for the second decoder $h^{s2}$. This cross-head mutual virtual adversarial training strategy incorporates complementary information and generates adversarial noises with reduced confirmation bias~\cite{arazo2020pseudo}. The training objective functions are as follows:

\begin{equation}
	\begin{aligned}
	\mathcal{L}_{\text {st1 }} = \sum_{x_i \in  D^{l}\cup  D^{u}} & \mathcal{D}\left( q^{t2}(y_i|x_i), p^{s1}(y_i|x_i+r^{adv}_{i1}, \theta^{s1})\right) \\
	&+\mathcal{D}\left(  q^{t1}(y_i|x_i), p^{s2}(y_i|x_i+r^{adv}_{i2}, \theta^{s2}) \right)
	\end{aligned}
        \label{loss:vammt}
\end{equation}

Where

\begin{equation}
\begin{array}{r}
r_{i 1}^{a d v} =\underset{\Delta r_1}{\arg \max } \mathcal{D} \left( q^{t2}(y_i|x_i), p^{s1}(y_i|x_i+\Delta r_1, \hat{\theta}^{s1})\right), \\ 
r_{i 2}^{a d v} =\underset{\Delta r_2}{\arg \max } \mathcal{D} \left( q^{t1}(y_i|x_i), p^{s2}(y_i|x_i+\Delta r_2, \hat{\theta}^{s2})\right),
\end{array}
\end{equation}

\begin{equation}
	q_{i}^{tm}=h^{tm}\left(g(x_{i})\right)
\end{equation}
where $p_i^{tm}$ denotes the prediction of m-th teacher segmentation head. Miyatoet al.~\cite{miyato2018virtual} provided a fast method to compute an approximation of $r_{i}^{a d v}$ , which was computed with one set of back-propagation of the network. For in-depth details, we referred the readers to~\cite{miyato2018virtual}. In addition, we select soft dice loss to measure the divergence:

\begin{equation}
	\mathcal{L}_{l d s} (q,p) =\frac{1}{C} \sum_{c=1}^C\left[1-\frac{2\left\|q\left(\hat{y}_c \mid x\right) \cap p\left(y_c \mid x+r^{a d v}\right)\right\|}{\left\|q\left(\hat{y}_c \mid x\right)\right\|+\left\|p\left(y_c \mid x+r^{a d v}\right)\right\|}\right]
\end{equation}
where $C$ is the number of classes. 

In  summary, MVAT seeks  the  direction  of  perturbation which  can  effectively  alter  the  distribution  at  the  data-level, thereby bolstering the model's ability to robustly extract discriminative information.

\subsection{Data argumentation with Cross-set CutMix} \label{CMMT-Mix}

Most existing semi-supervised learning strategy apply data augmentation on the unlabeled data to achieve consistency regularization. However, the discrepancy gaps between the labeled data and unlabeled data may decrease the performance of the models~\cite{wang2019semi}. To further reduce the domain shifts, following \cite{verma2022interpolation}, we expand the CutMix~\cite{yun2019cutmix} to combine both the labeled and unlabeled set. The key idea here is that we form new augmented samples by copying a local region of one image with a patch and paste them both on the labeled and unlabeled image, and the pixels are sampled from an estimated labeled data confidence distribution. The corresponding mixing procedure is:

\begin{equation}
	\begin{aligned}
	\tilde{x}= (1 - M) \odot x_a + M  \odot x_b \\
	\tilde{y}= (1 - M) \odot y_a + M  \odot y_b
	\end{aligned}
\end{equation}
where $M \in R^{H\times W\times C }$ is the binary mask, $x_a \in D^{u}$ and  $x_b \in  D^{l}\cup  D^{u}$.  This approach enables the generated CutMixed samples to incorporate the local region from both the labeled and unlabeled data, thereby reducing the discrepancy between these two datasets.

Similarly, after obtaining the CutMixed images $(\tilde{x}_i, \tilde{y}_i)$, we implement cross-head mutual mean-teaching strategies on them by adapting the mixed labels generated by the teacher to cross-supervise the student networks. Specifically, we utilize the mixed label $\tilde{y}_i^{t2}$ from the second teacher head $h^{t2}$ to guide the training of the student $f^{s1}$, and the output $\tilde{y}_i^{t1}$ from the first teacher head $h^{t1}$ to guide the training of the student $f^{s2}$. Then, we jointly train the cross-head teacher-student peer networks on the unlabeled data as follows:

\begin{equation}
	\mathcal{L}_{\text {st2}} =\sum_{x_i \in D^u}\mathcal{L}_{lds}\left(f_{s1}\left(\tilde{x}_{i}\right), \tilde{y}_i^{t2}\right)+\mathcal{L}_{\text {lds}}\left(f_{s2}\left(\tilde{x}_{i}\right), \tilde{y}_i^{t1}\right)
    \label{loss:cutmix}
\end{equation}

Where
\begin{equation}
	\tilde{y}_{i}^{tm}=(1 - M) \odot h^{tm}\left(g(x_{i})\right) + M \odot h^{tm}\left(g(x_{j}) \right), m \in \{ 1,2\}
\end{equation}

\subsection{Total Training Loss}

Finally, the overall training  objective is written as

\begin{equation}
	\mathcal{L}_{\text {all}}= \lambda \mathcal{L}_{\text {sup }}  + \alpha \mathcal{L}_{ss} + \beta (\mathcal{L}_{st1} + \mathcal{L}_{st2})
\end{equation}
Where $\lambda$, $\alpha$ and $\beta$ are the parameters to control the weight of the loss function. During the training phase, we set  $\lambda = 1.0 $, $\alpha = 2.0 $ and $\beta = 2.0$ for the 2D medical image segmentation tasks, $\lambda = 0.5 $, $\alpha = 0.5 $ and $\beta = 0.5$ for the 3D medical image segmentation tasks respectively.

\subsection{Training algorithm for CMMT-Net}

To elaborate on the training process of the proposed model, Algorithm~\ref{alg_TSML} summarizes the  detailed training procedure of our Cross-head Mutual Mean-Teaching Network (CMMT-Net).

\begin{algorithm}[!htbp]
	\renewcommand{\algorithmicrequire}{\textbf{Input:}}
	\renewcommand\algorithmicensure {\textbf{Output:} }
	\caption{ Semi-supervised medical image segmentation Cross-head Mutual Mean-Teaching Network (CMMT-Net).} \label{alg_TSML}
	\begin{algorithmic}[1]
		\REQUIRE ~~\\
		Labeled dataset: $D_l = \{(X_l, Y_l)\}_{i=1}^{N}$ ; \\
		Unlabeled dataset: $D_u=\{(X_u)\}_{j=1}^{M}$ ; \\
		Parameters: $\lambda$, $\alpha$, $\beta$; \\
		\ENSURE ~~\\
		Teacher model $f^{t}$ and student model $f^s$ .\\
            \STATE Initialization Student: Randomly initialize student model; \\
            Initialization Teacher: 
		\STATE Apply same initialization of student. \\
		\FOR{epoch $i \in [1,T]$}
            \STATE $x_i, y_i$=Sample($D^l$) \\
            \STATE $x_j$=Sample($D^u$) \\
		\STATE Perform Supervised Learning on the labeled data via Eq.~\ref{loss:sup};
		\STATE Perform Cross-head co-training with unlabeled data  via Eq.~\ref{loss:cct};
		\STATE Perform Mutual Virtual adversarial Training via Eq.~\ref{loss:vammt};
		\STATE Perform  Cross-head mutual mean-teaching with CutMix via Eq.~\ref{loss:cutmix}.
		\ENDFOR \\

	\end{algorithmic}
\end{algorithm}

\section{Experiments}

\begin{table*}[!h]
	\centering
	\setlength{\tabcolsep}{6pt}
	\caption{Comparisons with SOTA methods when using 10\%/20\% labeled cases on the LA dataset respectively. }
	\label{tab:la}
	\begin{tabular}{l|cc|ccccc}
		\toprule
		\multirow{2}{*}{Method} & \multicolumn{2}{c|}{Scans used}  & \multicolumn{4}{c}{Metrics}  \\ \cline{2-7}
		& Labeled & Unlabeled  & Dice(\%)$\uparrow $ & jaccard(\%)$\uparrow$ & $HD_{95}$(voxel)$\downarrow$  & ASD(voxel)$\downarrow$   \\
		\hline
		V-Net & 8 (10\%)   & 0  & 78.57 & 66.96 & 21.10 & 6.07  \\
		V-Net & 16 (20\%)  & 0  & 86.96 & 77.31 & 11.85 & 3.22  \\
		V-Net & 80 (100\%) & 0  & 91.62 & 84.60 & 5.40 & 1.64  \\
        V-Net (Dual Decoders) & 80 (100\%) & 0  & 92.11 & 85.43 & 5.29 & 1.43  \\
		\hline
		UA-MT~\cite{yu2019uncertainty} (MICCAI)  & \multirow{13}{*}{8 (10\%)} & \multirow{13}{*}{72(90\%) }& 86.28 & 76.11 & 18.71 & 4.63   \\
		SASSNet~\cite{li2020shape} (MICCAI) & & & 85.22 & 75.09 & 11.18 & 2.89\\
		DTC~\cite{luo2021semi} (AAAI) & & & 87.51 & 78.17 & 8.23 & 2.36  \\ 
		URPC~\cite{luo2021efficient} (MICCAI) & & & 85.01 & 74.36 & 15.37 & 3.96   \\
        SS-Net~\cite{wu2022exploring} & & & 88.55 & 79.62 & 7.49 & 1.90 \\
		MC-Net~\cite{wu2021semi} (MICCAI) & & & 87.50 & 77.98  & 11.28 & 2.30  \\
		MC-Net+~\cite{wu2022mutual} (MedIA 2022) & & & 88.96 & 80.25 & 7.93 & 1.86  \\
		CC-Net~\cite{huang2022complementary}   & & & 89.42 & 80.95 & 7.37 & 2.17  \\
        AC-MT~\cite{xu2023ambiguity} & & & 89.12 & 80.46  & 11.05  & 2.19  \\
        MLB-Seg~\cite{wei2023consistency} (MICCAI, 2023)  & & & 88.69 & 79.86 & 8.99 & 2.61  \\
        BCP~\cite{bai2023bidirectional} (CVPR, 2023) & & & 89.62 & 81.31 & 6.81 & 1.76  \\
        CAML~\cite{gao2023correlation} (MICCAI, 2023) ~  & & & 89.62 & 81.28 & 8.76 & 2.02  \\

		Ours & & & \textbf{90.75} & \textbf{83.13} & \textbf{5.77} & \textbf{1.82}  \\
		\hline
		UA-MT~\cite{yu2019uncertainty} (MICCAI) & \multirow{12}{*}{16 (20\%)} & \multirow{12}{*}{64 (80\%)} & 88.74 & 79.94 & 8.39 & 2.32 \\
		SASSNet~\cite{li2020shape} (MICCAI) & &  & 89.16 & 80.60 & 8.95 & 2.26   \\
		DTC~\cite{luo2021semi} (AAAI) & & & 89.52 & 81.22 & 7.07 & 1.96  \\
		URPC~\cite{luo2021efficient} (MICCAI) & & & 88.74 & 79.93 & 12.73 & 3.66    \\
		MC-Net~\cite{wu2021semi} (MICCAI) & & & 90.12 & 82.12  & 8.07 & 1.99  \\
		MC-Net+~\cite{wu2022mutual} (MIA) & & & 91.07 & 83.67 & 5.84 & 1.67  \\
        CC-Net~\cite{huang2022complementary}  (MIA)  & & & 91.14 & 83.79 & 5.74 & 1.57 \\
        MCF (CVPR, 2023)~\cite{wang2023mcf}  & & & 88.71 & 80.41 & 6.32 & 1.90  \\
        DC-Net~\cite{zhang2023cross} & & & 90.89 & 83.06 & 6.76 & 1.56   \\
        CAML~\cite{gao2023correlation} (MICCAI, 2023) ~  & & & 90.76 & 83.2 & 6.11 & 1.68  \\
        AC-MT~\cite{xu2023ambiguity}  (MedIA 2023) & & & 90.31 & 82.43 & 6.21 & 1.76 \\

		Ours & & & \textbf{91.83} & \textbf{84.95} & \textbf{4.95}  & \textbf{1.74}  \\
		\bottomrule
	\end{tabular}
\end{table*}

\subsection{ Experimental setup}

\subsubsection{ Datasets }

We comprehensively evaluate our method on three datasets: two 3D segmentation datasets  LA~\cite{LA} and  Pancreas-CT~\cite{Pancreas-CT} and a 2D segmentation dataset ACDC~\cite{ACDC}. The data processing follows MC-Net+~\cite{wu2022mutual}.

\textbf{LA} dataset is utilized as the benchmark for the 2018 Atrial Segmentation Challenge. This dataset consists of 100 gadolinium-enhanced MR imaging scans used for training, with an isotropic resolution of $0.625\times 0.625\times 0.625$ mm. Each image in the dataset is accompanied by segmentation masks for the left ventricle (LV), myocardium (Myo), and right ventricle (RV). In line with~\cite{luo2021semi}, consistent data partitioning is maintained across all experiments. This partitioning allocates 70 samples for training, 10 for validation, and 20 for testing.

\textbf{Pancreas-CT}  dataset is made public by the National Institutes of Health Clinical Center, which contains 82 3D abdominal contrast-enhanced CT scans collected with Philips and Siemens MDCT scanners, with a fixed in-plane resolution of 512 × 512 and varying intra-slice spacing from 1.5 to 2.5 mm. The data split is fixed with previous works~\cite{luo2022semi}. 62 samples are used for training, and performance is reported on the rest 20 samples. For necessary preprocessing, the Hounsfield Units (HU) in all the CT scans have been rescaled, with a window level of 75 and a window width of 400. Then we resample all the scans into an isotropic resolution of 1.0 mm × 1.0 mm × 1.0 mm. We apply the same settings (i.e. training with 10\% labeled data and training with 20\% labeled data) as LA dataset in the experiments.  

\textbf{ACDC} dataset  comprises 200 annotated short-axis cardiac MR-cine images from 100 patients. Each image is provided with segmentation masks for the left ventricle (LV), myocardium (Myo), and right ventricle (RV). Consistent with~\cite{luo2021semi}, a uniform data partitioning scheme is employed for all experiments, designating 70 samples for training, 10 for validation, and 20 for testing. All slices are resized to $256\times 256$ to train the models.

\begin{table*}[!htbp]
	\centering
	\setlength{\tabcolsep}{6pt}
	\caption{Comparisons with SOTA methods when using 10\%/20\% labeled cases on the Pancreas-CT dataset respectively.}
	\label{tab:pancreas-ct}
	\begin{tabular}{l|cc|cccc}
		\toprule
		\multirow{2}{*}{Method} & \multicolumn{2}{c|}{Scans used}  & \multicolumn{4}{c}{Metrics}  \\ \cline{2-7}
		& Labeled & Unlabeled  & Dice (\%)$\uparrow $ &                
            jaccard (\%)$\uparrow$ & $HD_{95}$(voxel)$\downarrow$  &        
            ASD (voxel)$\downarrow$   \\
		\hline
		V-Net & 6(10\%)   & 0 & 54.94 & 40.87 & 47.48 & 17.43 \\
		V-Net & 12(20\%)  & 0 & 71.52 & 57.68 & 18.12 & 5.41 \\
		V-Net & 62(100\%) & 0 & 82.60 & 70.81 & 5.61 & 1.33  \\
        V-Net (Daul Decoder) & 62(100\%) & 0 & 83.31 & 71.77 & 5.11 & 1.27  \\
		\hline
		UA-MT~\cite{yu2019uncertainty} (MICCAI) & \multirow{9}{*}{6 (10\%)} & \multirow{9}{*}{56 (90\%)} & 66.44 & 52.02 & 17.04 & 3.03   \\
		SASSNet~\cite{li2020shape} (MICCAI) & &  & 68.97 & 54.29 & 18.83 & 1.96 \\
		DTC~\cite{luo2021semi} (AAAI) & & & 66.58 & 51.79 & 15.46 & 4.16  \\
		URPC~\cite{luo2021efficient} (MICCAI) & & & 73.53 & 59.44 & 22.57 & 7.85    \\
		MC-Net~\cite{wu2021semi} (MICCAI) & & & 69.07 & 54.36  & 14.53 & 2.28  \\
		MC-Net+~\cite{wu2022mutual} (MIA) & & & 70.00 & 55.66  & 16.03 & 3.87 \\
		Multi-scale MC-Net+~\cite{wu2022mutual} (MIA) & & & 74.01 & 60.02 & 12.59 & 3.84  \\
		Ours & & & \textbf{82.73} & \textbf{70.83} & \textbf{5.38} & \textbf{1.90} \\
		\hline
		UA-MT~\cite{yu2019uncertainty} (MICCAI) & \multirow{10}{*}{12 (20\%)}  & \multirow{10}{*}{50(80\%)} & 76.01 & 62.62 & 10.84 & 2.43   \\
		SASSNet~\cite{li2020shape} (MICCAI ) & &  & 76.39 & 62.62 & 12.59 & 3.34  \\
		DTC~\cite{luo2021semi} (AAAI) & & & 76.27 & 62.82 & 8.70 & 2.20  \\
		URPC~\cite{luo2021efficient} (MICCAI) & & & 80.02 & 67.30 & 8.51 & 1.98    \\
		MC-Net~\cite{wu2021semi} (MICCAI) & & & 78.17 & 65.22  & 6.90 & 1.55 \\
		MC-Net+~\cite{wu2022mutual} (MIA) & & & 79.37 & 66.83 & 8.52 & 1.72  \\
		Multi-scale MC-Net+~\cite{wu2022mutual} (MIA) & & & 80.59 & 68.08 & 6.47 & 1.74  \\
        BCP~\cite{bai2023bidirectional} (CVPR 2023) & & & 82.91 & 70.97 & 6.43 & 2.25 \\
        DC-Net~\cite{zhang2023cross} & & & 81.32 & 68.45 & 1.20 & 6.53  \\

  		Ours & & & \textbf{83.48} & \textbf{71.35} & \textbf{5.36} & \textbf{1.80} \\
		\bottomrule
	\end{tabular}
\end{table*}

\subsubsection{ Experimental details }


\textbf{3D Segmentation:} We carried out a normalization process to obtain zero mean and unit variance before entering the 3D images into the networks. Following the approach outlined in MC-Net+~\cite{wu2022mutual}, using expanded  margins of [10 $\sim$ 20, 10 $\sim$ 20, 5 $\sim$ 10] or [25, 25, 0] voxels for LA or Pancreas-CT, respectively. In the training phase, we randomly extracted 3D patches with sizes of 96 x 96 x 96 for Pancreas-CT or 112 x 112 x 80 for LA. Additionally, we used 2D rotation and flip operations for data augmentation on the LA dataset. A batch size of four was used for both datasets, with each batch including two labeled patches and two unlabeled patches. Our 3D CMMT-Net model adopted the V-Net architecture as its backbone,  and was trained for a duration of 15k iterations. During testing, we adopted a sliding window of size 112 $\times$ 112 $\times$ 80 or 96 $\times$ 96 $\times$ 96, with a fixed stride of 18 $\times$ 18 $\times$ 4 or 16 $\times$ 16 $\times$ 16 for LA or Pancreas-CT, respectively.  We then combined the predictions obtained from the patch-based approach to derive the final comprehensive results.

\textbf{2D Segmentation:} In preparation for the ACDC dataset, we first performed sample normalization to achieve a zero mean and unit variance. Data augmentation was implemented through random rotations and flips. We extracted 2D patches, each sized at $256 \times 256$, in a random manner, utilizing a batch size of 24. Each batch was composed of 12 labeled data samples and 12 unlabeled samples. During the testing phase, we initially resized the scans to dimensions of $256 \times 256$ for input processing and later restored them to their original size to obtain the final results.  Our 2D CMMT-Net model adopted the U-Net architecture as its backbone, and was  trained for a total of 30,000 iterations. All experimental configurations for the ACDC dataset strictly adhered to the publicly established benchmark~\cite{wu2022mutual} to ensure fair and consistent comparisons.

We used the SGD optimizer with a learning rate of  $10^{-2}$ and a weight decay factor of $10^{-4}$ for training on all datasets.  For the 2D segmentation task, the parameters are set to $\lambda=1.0$, $\alpha=2.0$, and $\beta=2.0$, while for the 3D segmentation task, they are set to $\lambda=0.5$, $\alpha=0.4$, and $\beta=0.5$. The weights  $\lambda$, $\alpha$ and $\beta$ were set using a time-dependent Gaussian warming-up function. We carried out two standard semi-supervised experimental setups, i.e., training with 10\% or 20\% labeled data with the remaining data being unlabeled. For the quantitative assessment, we used four metrics: Dice, Jaccard, the average surface distance (ASD), and the 95 percent Hausdorff distance ($HD_{95}$).

\subsubsection{Compared methods}

\textbf{Baselines and Competitors.} To simply illustrate the effectiveness of our CMMT-Net, we have created the following benchmark techniques and rivals: (1)~\textbf{Learning with Limited Data (LS)}: This approach is trained solely with labeled data, leaving out unlabeled data entirely. It represents the lower performance bound. (2)~\textbf{Fully Supervised Learning (FL)}: Training is carried out with the aid of all available data, demonstrating the upper bound on performance. (3)~We compare our methods with various SOTA SSMIS approaches, including: uncertainty-aware mean teacher (UA-MT)~\cite{yu2019uncertainty}, shape-aware semi-supervised net (SASSnet)~\cite{li2020shape}, dual-task consistency (DTC)~\cite{luo2021semi}, Uncertainty Rectified Pyramid Consistency (URPC)~\cite{luo2021efficient}, mutual consistency network (MC-Net~\cite{wu2021semi}), MC-Net+~\cite{wu2022mutual}, and CC-Net~\cite{huang2022complementary}, SS-Net~\cite{wu2022exploring}, AC-MT~\cite{xu2023ambiguity}, MLB-Seg~\cite{wei2023consistency}, BCP~\cite{bai2023bidirectional}, CAML~\cite{gao2023correlation}, AC-MT~\cite{xu2023ambiguity},  MCF~\cite{wang2023mcf}, CAML~\cite{gao2023correlation}, SASSNet~\cite{li2020shape}, DC-Net~\cite{zhang2023cross}.

\subsection{Quantitative results}

\subsubsection{Performance on the LA dataset}

Table \ref{tab:la} reports the results using 10\% and 20\% of the labeled data for training on the LA dataset as mentioned above. Comparing with the baseline V-Net, our CMMT-Net trained with only 10\% labeled data achieved 90.75\%, 83.19\%, 5.77 voxels, and 1.82 voxels on the four evaluation metrics of Dice, Jaccard, $HD_{95}$, and ASD, respectively. While using 20\% labeled data for training,  our CMMT-Net achieves a remarkable Dice performance of 91.83\%,  surpassing the fully supervised V-Net by 0.21\% (91.83\% vs. 91.62\%) and  falling only 0.28\% behind the fully supervised V-Net (Dual Decoders), indicating the effectiveness of our Cross-head mutual mean-teaching method. Taking a closer look at the results, with only 10\% labeled data for training, compared to MC-Net+~\cite{CC-Net} which adopts a shared encoder with  three slightly different decoders for SSMIS tasks, our method obtains a gain of 1.79\% (90.75\% vs. 88.96\%) Dice performance. As for CC-Net~\cite{huang2022complementary}, which incorporates two additional complementary auxiliary networks to achieve inter-model perturbations and enforce consistency among these three models, the Dice performance improved from 89.42\% to 90.75\%. Compared to AC-MT~\cite{xu2023ambiguity} which adopts teacher-student framework to achieve ambiguity-selective consistency learning, our CMMT-Net gains 1.63\% (90.75\% vs. 89.12\%) in terms of Dice performance. Unlike these methods only explore the perturbations on the feature level or network level, our CMMT-Net further incorporates with weak and strong augmentation, thereby demonstrating the effectiveness of our data-level perturbations. Besides, Compared to BCP~\cite{bai2023bidirectional} which joint processes the labeled and unlabeled data using a copy-paste strategy and encourages the prediction consistency between the teacher model and the student model,  there is an improvement of 1.13\% (90.75\% vs. 89.62\%), indicating the effectiveness of our dual decoder strategy. When trained with 20\% labeled data, it is obvious that our CMMT-Net outperforms all the other SOTA methods, demonstrating the superiority of our method.

\begin{table*}[!h]
	\centering
	\setlength{\tabcolsep}{6pt}
	\caption{Comparisons with SOTA methods when using 10\%/20\% labeled cases on the ACDC dataset respectively. }
	\label{tab:acdc}
	\begin{tabular}{l|cc|cccc}
		\toprule
		\multirow{2}{*}{Method} & \multicolumn{2}{c|}{Scans used}  & \multicolumn{4}{c}{Metrics}  \\ \cline{2-7}
		& Labeled & Unlabeled  & Dice(\%)$\uparrow $ &                
            jaccard(\%)$\uparrow$ & $HD_{95}$(voxel)$\downarrow$  &        
            ASD(voxel)$\downarrow$   \\     
		\hline
		Unet & 7 (10\%) & 0  & 77.34 & 66.20 & 9.18 & 2.45  \\
		Unet & 14 (20\%) & 0 & 85.15 & 75.48 & 6.20 & 2.12  \\
		Unet & 70 (100\%) & 0 & 91.65 & 84.93 & 1.89 & 0.56 \\
		\hline
		Ours & & & \textbf{87.13} & \textbf{78.01} & \textbf{2.30} & \textbf{0.62}   \\
		\hline
		UA-MT~\cite{yu2019uncertainty} (MICCAI) & \multirow{9}{*}{7 (10\%)} &  \multirow{9}{*}{63 (90\%)}  & 81.58 & 70.48 & 12.35 & 3.62   \\
		SASSNet~\cite{li2020shape} (MICCAI) &  & & 84.14 & 74.09 & 5.03 & 1.40 \\
		DTC~\cite{luo2021semi} (AAAI) & &  & 82.71 & 72.14 & 11.31 & 2.99 \\
		URPC~\cite{luo2021efficient} (MICCAI) & & & 81.77 & 70.85 & 5.04 & 1.41    \\
		MC-Net~\cite{wu2021semi} (MICCAI) & & & 86.34 & 76.82  & 7.08 & 2.08  \\
		MC-Net+~\cite{wu2022mutual} (MIA) & & & 87.10 & 78.06 & 6.68 & 2.00  \\
        SS-Net~\cite{wu2022exploring} & & & 86.78 & 77.67 & 6.07 & 1.40 \\
        BCP~\cite{bai2023bidirectional} (CVPR 2023) & & & 88.84 & 80.62 & 3.98 & 1.17 \\
		Ours  & & & \textbf{90.67} & \textbf{83.43} & \textbf{1.33} & \textbf{0.34} \\
		\hline
		UA-MT~\cite{yu2019uncertainty} (MICCAI) & \multirow{8}{*}{14 (20\%)} & \multirow{8}{*}{56 (80\%)}  & 85.87 & 76.78 & 6.68 & 2.00 \\
		SASSNet~\cite{li2020shape} (MICCAI) & &  & 87.04 & 78.13 & 7.84 & 2.15  \\
		DTC~\cite{luo2021semi} (AAAI) & & & 86.28 & 77.03 & 6.14 & 2.11 \\
		URPC~\cite{luo2021efficient} (MICCAI) & & & 85.07 & 75.61 & 6.26 & 2.11   \\
		MC-Net~\cite{wu2021semi} (MICCAI) & & & 87.83 & 79.14  & 4.94 & 1.52  \\
		MC-Net+~\cite{wu2022mutual} (MIA) & & & 88.51 & 80.19 & 5.35 & 1.54  \\
        FBA-Net~\cite{chung2023fba} & & & 89.81 & - & - & 1.11 \\
        Ours & & & \textbf{91.11} & \textbf{84.10} & \textbf{1.17} & \textbf{0.29}  \\
		\bottomrule
	\end{tabular}
\end{table*}

\subsubsection{Performance on the pancreas-CT dataset}

Table~\ref{tab:pancreas-ct} illustrates the results obtained by our CMMT-Net and other comparative SSMIS methods on the Pancreas-CT dataset. Our proposed CMMT-Net stands out with the highest Dice and Jaccard scores, underscoring its effectiveness. With just 10\% labeled data for training, our model achieves a Dice score of 82.73\%, while MC-Net+ lags significantly at 70.00\%, representing a substantial improvement of 12.73\%. Even with a higher proportion of labeled data (20\%), our CMMT-Net still outperforms MC-Net+ by 4.11\% (82.73\% vs. 79.37\%). These results not only highlight the superior performance of our method but also emphasize its robustness across varying scenarios. When compared to BCP~\cite{bai2023bidirectional}, which employs a copy-paste strategy for augmenting both labeled and unlabeled data, our CMMT-Net exhibits a slight improvement of 0.57\% (83.48\% vs. 82.91\%) in Dice performance, underscoring the significance of data-level augmentation. Remarkably, with just 20\% labeled data for training, our CMMT-Net even surpasses the fully-supervised V-Net (Dual decoder) in terms of dice performance. This remarkable achievement can be attributed to the sophisticated architecture of our CMMT-Net and the extensive application of data augmentation to both labeled and unlabeled data. The systematic combination of three levels of perturbations empowers our model to achieve higher accuracy and robust performance, even outperforming fully-supervised learning methods.

\subsubsection{Performance on the ACDC dataset}

Table~\ref{tab:acdc}  presents quantitative results on the ACDC dataset with varying proportions of labeled training data: 10\%, and 20\%. Our CMMT-Net  consistently outperforms other SOTA methods across diverse semi-supervised learning scenarios, exhibiting a significant advantage in dice metrics and securing the best performance in terms of all the val metrics. Specifically, when utilizing 10\% labeled data for training,  compared with the previously best-performing method BCP~\cite{bai2023bidirectional}, our CMMT-Net yields improvements of 1.83\% (90.67 \% 88.84), indicating the effectiveness of our method.

\begin{table}[!h]
	\centering
	\begin{threeparttable}
	\setlength{\tabcolsep}{4pt}
	\setlength{\belowcaptionskip}{10pt}
	\caption{Ablation studies of our CMMT-Net on the LA dataset. Note that, CCT means using Cross-head Co-Training strategy; MT means using mean teacher strategy, i.e., adopting the predictions of the $h^{t1}$ to supervise the training of $h^{s1}$ and the predictions of the $h^{t2}$ to supervise the training of the $h^{s2}$;  MMT means using mutual mean teacher strategy, i.e., adopting the predictions of the $h^{t2}$ to supervise the training of the $h^{s1}$ and the predictions of the $h^{t2}$ to supervise the training of the $h^{s1}$; VAT means using  Virtual Adversarial Training strategy, CutMix  means using  CutMix augmentation to perturb the data.}
	\label{Table:component_analysis_la}
	\begin{tabular}{c|ccccc|c}
		\toprule
		\multirow{2}{*}{Labeled} &\multirow{2}{*}{ CCT} & \multirow{2}{*}{MT}  & \multirow{2}{*}{MMT}   & \multirow{2}{*}{VAT} & \multirow{2}{*}{Cutmix} &  \multicolumn{1}{l}{Metrics}   \\
		\cline{7-7}
		&  & & & & &  Dice(\%)$\uparrow $    \\
		\hline
		\multirow{12}{*}{8/72 (10\%)} & $\surd$ & & &  &  & 89.62   \\
		& &  & & $\surd$  & $\surd$  & 89.84  \\
		& $\surd$ & & &  $\surd$  & & 90.08   \\
		& $\surd$  &  & &  & $\surd$ & 89.55  \\
		& $\surd $ &  & & $\surd$ & $\surd$ & 89.82 \\  
            \cline{2-7}
		& $\surd $ & $\surd $ &  & $\surd$  &   & 89.70  \\
		& $\surd $ & $\surd $ &  &  & $\surd$   & 90.56 \\
            & $\surd $ & $\surd $ &  & $\surd$  & $\surd$  & 90.39  \\ 
            \cline{2-7}
            & $\surd $ &  &  $\surd $ & $\surd $ &  & 89.65  \\
            & $\surd $ &  &  $\surd $ &  & $\surd $ & 90.32  \\
            & &  &  $\surd $ & $\surd $ & $\surd $ & 90.60  \\
		& $\surd $ &  &  $\surd $ & $\surd $ & $\surd $ & \textbf{90.75}  \\
		\hline
            \hline		\multirow{12}{*}{16/64 (20\%)} &  $\surd$  & & &  &  & 91.18   \\
		&  & & & $\surd$  & $\surd$  &   90.97  \\
		& $\surd$ &  & & $\surd$ & & 90.89   \\
		& $\surd$  &   & & & $\surd$ & 90.72  \\
  		& $\surd$  &   & & $\surd$  & $\surd$ & 90.94  \\   
            \cline{2-7}
		& $\surd $ & $\surd $ &  & $\surd$ &  & 91.19  \\
		& $\surd $ & $\surd$ &  &  & $\surd$ & 91.53  \\
		& $\surd $ & $\surd $ &  & $\surd$  & $\surd $  & 91.49 \\ 
            \cline{2-7}
        & $\surd $ &  &  $\surd $ & $\surd $ & & 91.10   \\
		& $\surd $ &  &  $\surd $ &  & $\surd $ & 91.60   \\  
        &  &  & $\surd $ & $\surd$  & $\surd $  & 91.43 \\
		& $\surd $ &  &  $\surd $ & $\surd $ & $\surd $ & \textbf{91.83}   \\
		\bottomrule
	\end{tabular}
\end{threeparttable}
\end{table}

In summary, with a shared encoder and  only two different segmentation heads, our CMMT-Net yields the best performance and surpasses the MC-Net+~\cite{wu2022mutual} which design three various branches and  BCP~\cite{bai2023bidirectional} which adopts strong augmentation on the unlabeled data with a large margin in terms of the Dice performance, demonstrating the effectiveness of our cross-head mutual mean-teaching approach. 
More specifically, Our approach unifies data, feature, and network perturbations into a cohesive framework, and the synergy between these perturbations significantly contributes to our model's superior performance.  Unlike previous methods that tend to adopt increasingly complex designs, such as URPC~\cite{luo2021efficient}, MC-Net+\cite{wu2022mutual}, SS-Net\cite{wu2022exploring}, and DC-Net~\cite{zhang2023cross},  we propose a simple yet highly effective approach that underscores the importance of data perturbation and model stabilization in enhancing SSMIS performance across different labeling scenarios and diverse datasets.

\subsection{Analysis}

\subsubsection{Ablation study on different losses.}

Each component of our CMMT-Net contributes differently to the generalization of the models, and the combinations of these strategies yield significant  improvements compared with previous works. We  conducted experiments on two 3D segmentation dataset LA  and Pancreas-CT  with 10\% and 20\% labeled dataset respectively, as is illustrated in Tables~\ref{Table:component_analysis_la} and ~\ref{Table:component_analysis_pct}.

\textbf{Effectiveness of mutual mean-teaching.} To introduce network-level perturbations, we incorporate the mean-teacher (MT) strategy~\cite{tarvainen2017mean} by performing an exponential moving average of the student network's parameters, thereby further reducing the uncertainty and increasing the robustness of the model. However, directly applying the original MT strategy may not always lead to performance improvement, as illustrated in Tables~\ref{Table:component_analysis_la} and ~\ref{Table:component_analysis_pct}.  A closer analysis of Table~\ref{Table:component_analysis_la} reveals that, when training with a 10\% labeled dataset,  combining CCT and VAT, further adopting the MT strategy resulted in a performance drop from 90.08\% to 89.7\%. However, when combining CCT and CutMix, further incorporating the MT strategy led to an improvement in dice performance from 89.55\% to 90.56\%.  Based on the combination of CCT, VAT and CutMix, when incorporating MMT instead of  MT, our final performance reached 90.75\% and yields an improvement of 0.36\% (90.75\%  vs. 90.39\%), indicating the effectiveness of our mutual mean-teaching strategy. As observed in Table~\ref{Table:component_analysis_pct}, the dice performance improved by 1.53\% (82.73\% vs. 81.20\%) with 10\% labeled dataset for training and 1.34\% (83.48\% vs. 82.14) with a 20\% labeled dataset for training by replacing the MT with MMT strategy. Note that the combination of MMT, VAT, and CutMix even surpasses the performance of the combination of CCT, MMT, VAT, and CutMix. The reason behind this could be that cross-head co-training with hard pseudo-labels may introduce inevitable noise, which could potentially harm the performance of the models.

\begin{table}[!htbp]
	\centering
	\begin{threeparttable}
	\setlength{\tabcolsep}{4pt}
	\setlength{\belowcaptionskip}{10pt}
	\caption{Ablation studies of our CMMT-Net on the Pancreas-CT  dataset. The meanings of CCT, MT, MMT, VAT and CutMix are explained in Table~\ref{Table:component_analysis_la}.}
	\label{Table:component_analysis_pct}
	\begin{tabular}{c|ccccc|c}
		\toprule
		\multirow{2}{*}{Labeled} &\multirow{2}{*}{ CCT} & \multirow{2}{*}{MT}  & \multirow{2}{*}{MMT}   & \multirow{2}{*}{VAT} & \multirow{2}{*}{Cutmix} &  \multicolumn{1}{l}{Metrics}   \\
		\cline{7-7}
		&  & & & & &  Dice(\%)$\uparrow $    \\
		\hline
		\multirow{6}{*}{6/56 (10\%)} & $\surd$ & & &  &  & 79.14   \\
            & $\surd $ & $\surd $ &  & $\surd$  & $\surd$  & 81.20  \\ 
            & $\surd $ &  &  $\surd $ & $\surd $ &  & 82.59  \\
            & $\surd $ &  &  $\surd $ &  & $\surd $ & 79.78  \\
            & &  &  $\surd $ & $\surd $ & $\surd $ & 82.21  \\
		& $\surd $ &  &  $\surd $ &    $\surd $ & $\surd $ & \textbf{82.73} \\
		\hline
            \hline
		\multirow{6}{*}{12/50 (20\%)} &  $\surd$  & & &  &  & 80.93   \\
		& $\surd $ & $\surd $ &  & $\surd$  & $\surd $  & 82.14 \\ 
  	& $\surd $ &  &  $\surd $ & $\surd $ & & 80.64   \\
		& $\surd $ &  &  $\surd $ &  & $\surd $ & 83.32   \\   
        &  &  & $\surd $ & $\surd$  & $\surd $  & \textbf{83.69} \\
		& $\surd $ &  &  $\surd $ & $\surd $ & $\surd $ & 83.48   \\
		\bottomrule
	\end{tabular}
\end{threeparttable}
\end{table}

\begin{figure*}[!htbp]
	\centering

	\subfigure[Sensitivity to $\beta$]{
		\begin{minipage}[t]{0.48\linewidth}
			\centering
			\includegraphics[width=8.0cm]{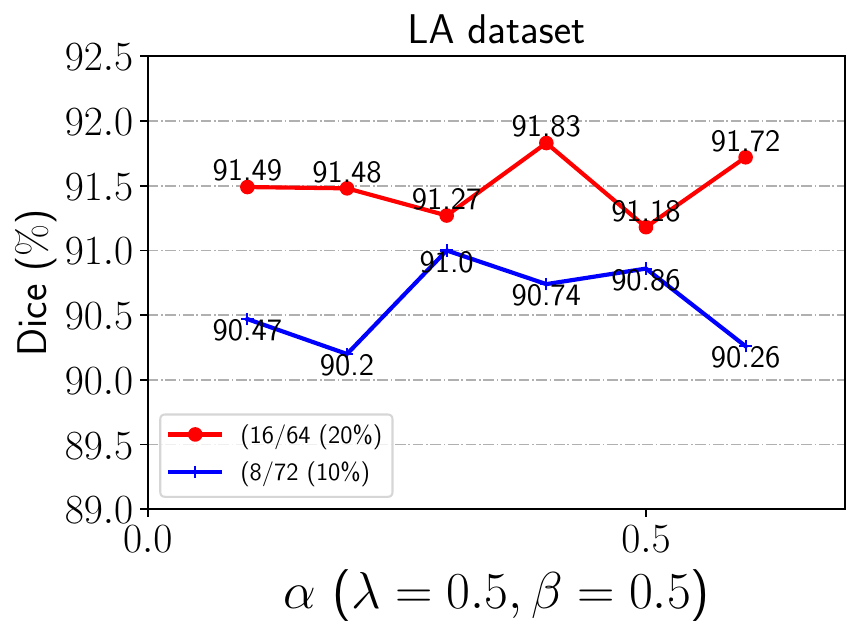}
		\end{minipage}
	}
	\subfigure[Sensitivity to $\tau$]{
		\begin{minipage}[t]{0.48\linewidth}
			\centering
			\includegraphics[width=8.0cm]{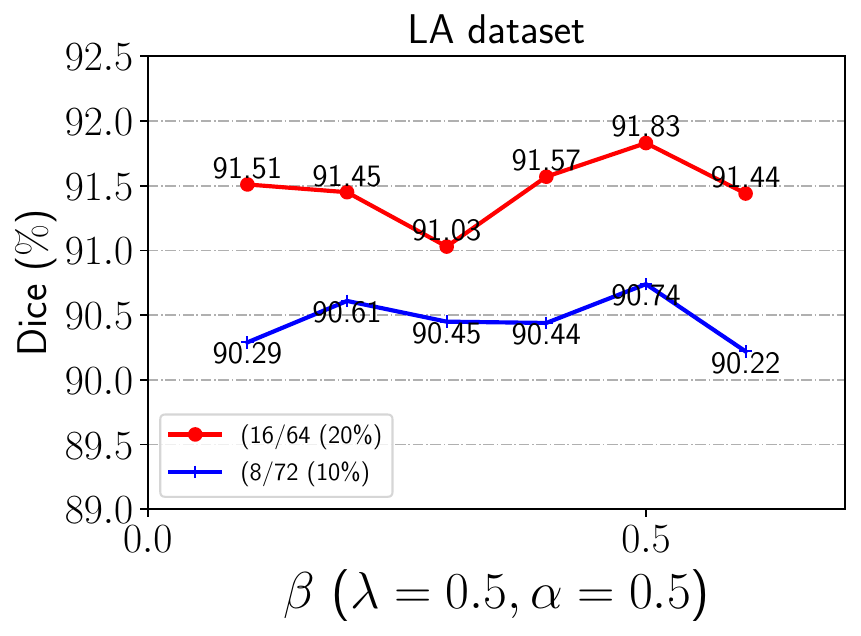}
		\end{minipage}
	}
	\caption{Performance sensitivity of the parameters $\alpha$ and  $\beta$  with 10\% labeled data on LA dataset.}
	\label{fig:sensitivity}
\end{figure*}

\textbf{Effectiveness of Mutual Virtual adversarial Training.} To introduce data perturbations and achieve a locally smooth decision boundary, we employ Mutual Virtual Adversarial Training (MVAT) for data augmentation. In Table~\ref{Table:component_analysis_la,} it is apparent that VAT does not consistently result in improved model performance. For instance, with only 10\% labeled data used for training, introducing VAT into the combination of CCT, MT, and CutMix actually leads to a decrease in performance, dropping from 90.56\% to 90.39\%. However, when incorporated with other methods, performance can be further enhanced. Specifically, by adopting VAT, the Dice performance surpasses the baseline CCT by 0.46\% (90.08\% vs. 89.62\%). In contrast, when combined with MT or MMT, the improvement is only 0.08\% (89.70\% vs. 89.62\%) and 0.03\% (89.65\% vs. 89.62\%), respectively. The further incorporation of CutMix results in our CMMT-Net achieving 90.75\% Dice performance, outperforming the baseline by 0.33\% (90.75\% vs. 90.32\%). Similar conclusions can be drawn from Table~\ref{Table:component_analysis_pct}.

\textbf{Effectiveness of Cross-set CutMix.} To introduce data perturbations and mitigate the distribution disparity between the labeled and unlabeled data, we employ Cross-set CutMix, a method that generates new augmented samples. This is achieved by selecting a local region from both labeled and unlabeled data, extracting a patch, and then pasting it onto the  image. Our findings, as depicted in Tables~\ref{Table:component_analysis_la} and \ref{Table:component_analysis_pct}, illustrate that the efficacy of our Cross-set CutMix strategy may vary when combined with different techniques. In particular, on Table~\ref{Table:component_analysis_la}, when utilizing a 10\% labeled dataset for training, the incorporation of Cross-set CutMix results in a minor 0.07\% reduction in Dice performance compared to the baseline CCT (89.55\% vs. 89.62\%). However, when used in conjunction with MT or MMT methods, it leads to notable improvements in Dice performance. In the case of Table~\ref{Table:component_analysis_pct}, with a 10\% labeled dataset for training, Cross-Set CutMix outperforms the baseline CCT by a mere 0.64\% (79.78\% vs. 79.14\%) when combined with MMT. In contrast, the improvement surges to 2.93\% (82.73\% vs. 79.14\%) when integrating all these strategies. Interestingly, VAT tends to perform better when the available labeled data for training is limited. With 10\% labeled data for training, the combination of CCT, MMT, VAT outperforms the combination of CCT, MMt, CutMix by 2.81\% (82.59\% vs 79.78\%), while drops 2.64\%  (80.64\% vs. 83.32\%) when utilizing 20\% labeled data for training. This suggests that the effectiveness of different strategies may vary under different conditions, but their amalgamation can yield better performance. Furthermore, unlike conventional methods that solely apply the CutMix strategy to unlabeled data, we extend it to labeled data as well. This is driven by the observation that labeled and unlabeled data exhibit domain shifts to some extent. By blending these two datasets, we can mitigate domain discrepancies, thereby enhancing the generalization of the model.

\subsubsection{Parameter sensitivity}

The parameters $\alpha$ and $\beta$ are essential to control the importance of each loss in the objective function.  To assess the sensitivity of these parameters, we conducted experiments using 10\% and 20\% of labeled data from the LA dataset for training. Specifically, $\alpha$ governs the adjustments applied to CCT, while $\beta$ influences the adjustments related to data perturbation. In Figure~\ref{fig:sensitivity}, we present the results for various parameter settings, keeping one parameter constant while varying the other. It can be  observed that our CMMT-Net achieves its best performance of 91.83\% with 20\% labeled data for training when $\alpha=0.4$ and $\beta=0.5$. Notably, with a 10\% labeled dataset for training, setting $\alpha=0.3$ and $\beta=0.5$ results in our CMMT-Net achieving its best performance of 91.0\%, surpassing the performance of 90.74\% by 0.26\%. It is important to note that a smaller $\alpha$ can lead to reduced performance due to inadequate training on unlabeled data, while a larger $\alpha$ can produce inaccurate results, also resulting in sub-optimal performance. Similarly, both smaller and larger values of $\beta$ can lead to sub-optimal performance. Therefore, in this study, we set $\alpha=0.4$ and $\beta=0.5$ to balance the two losses for both the LA and Pancreas-CT datasets. Furthermore, the consistency of dice values across different $\alpha$ and $\beta$ values in each semi-supervised setting indicates the robustness of our model to these hyperparameters. Additionally, the results exhibit minimal fluctuations, further emphasizing the robustness of our approach.

\begin{figure*}[!h]
	 \centering
	\includegraphics[scale=0.17]{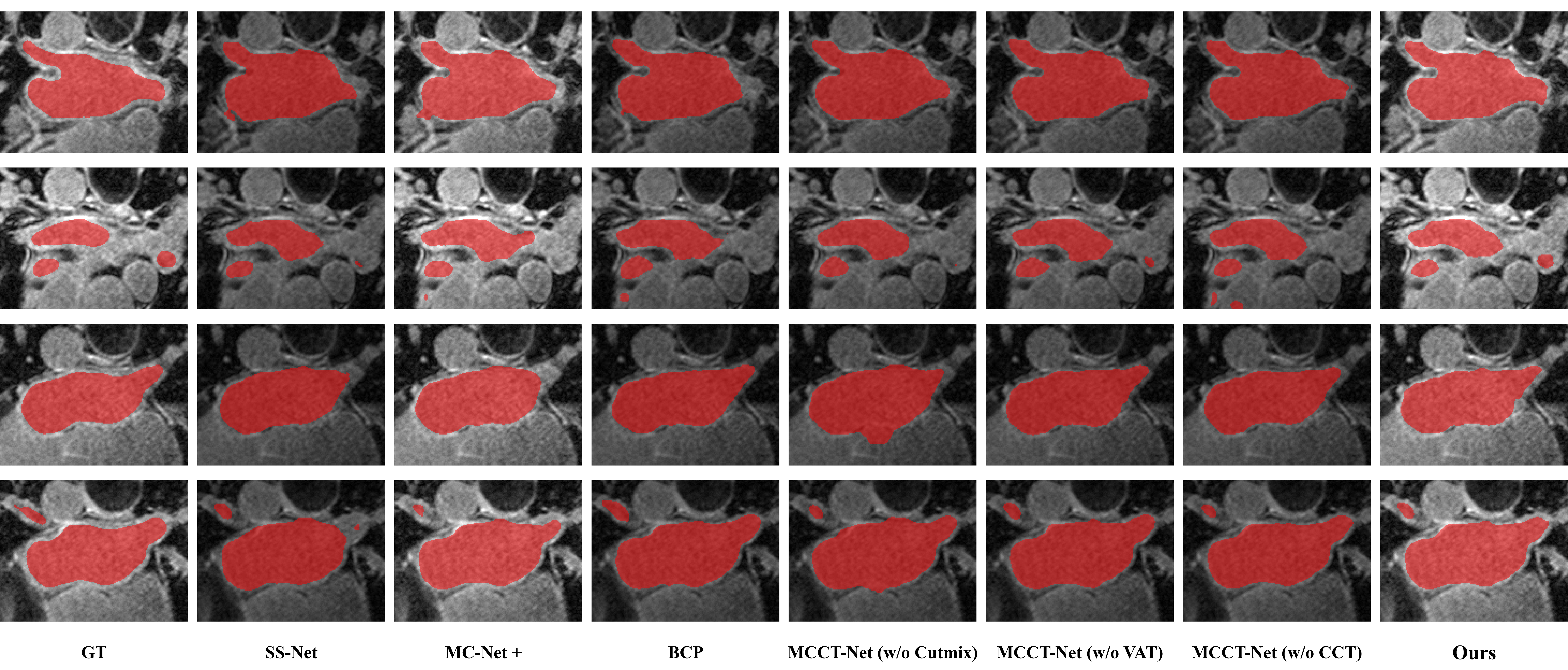}
	\caption{\textbf{Visualizations of several semi-supervised segmentation methods with 10\% labeled data and ground truth on LA dataset}.}
	\label{fig:la}
\end{figure*}

\begin{figure*}[!h]
	\centering
     \includegraphics[scale=0.035]{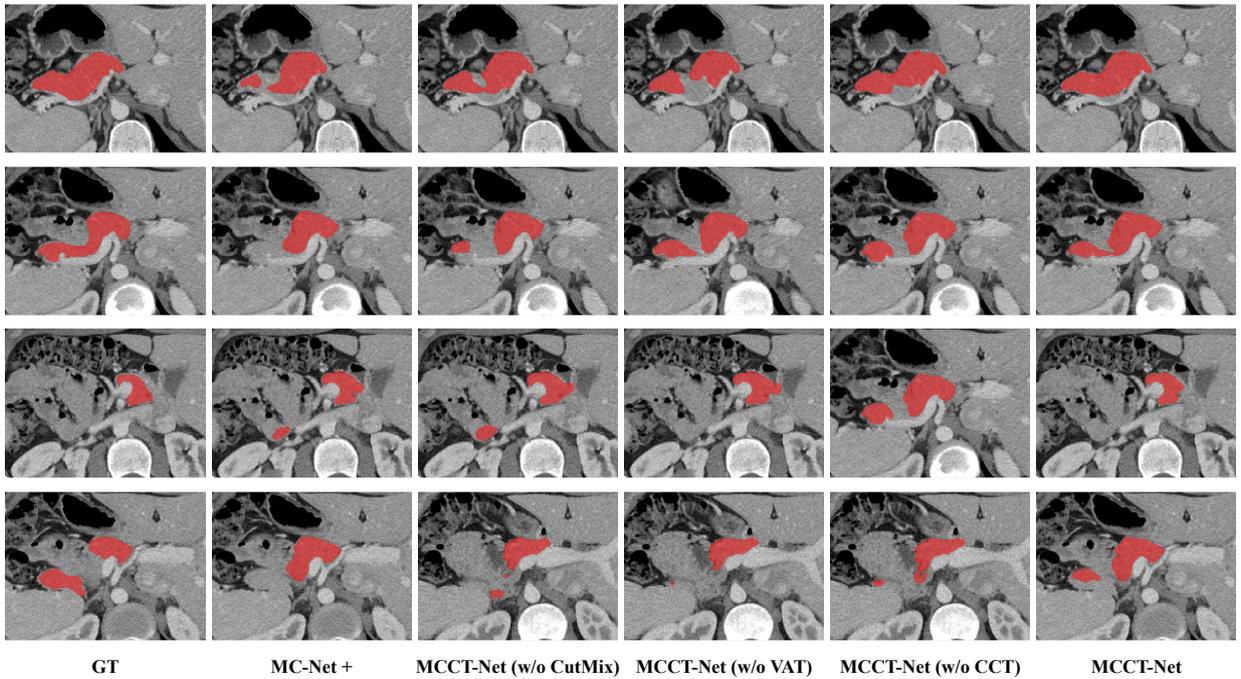}
	\caption{\textbf{Visualizations of several semi-supervised segmentation methods with 10\% labeled data and ground truth on Pancreas CT dataset}.}
	\label{fig:pct}
\end{figure*}

\subsubsection{Qualitative Visualization}

\begin{figure*}[!h]
	\centering
	\includegraphics[scale=0.08]{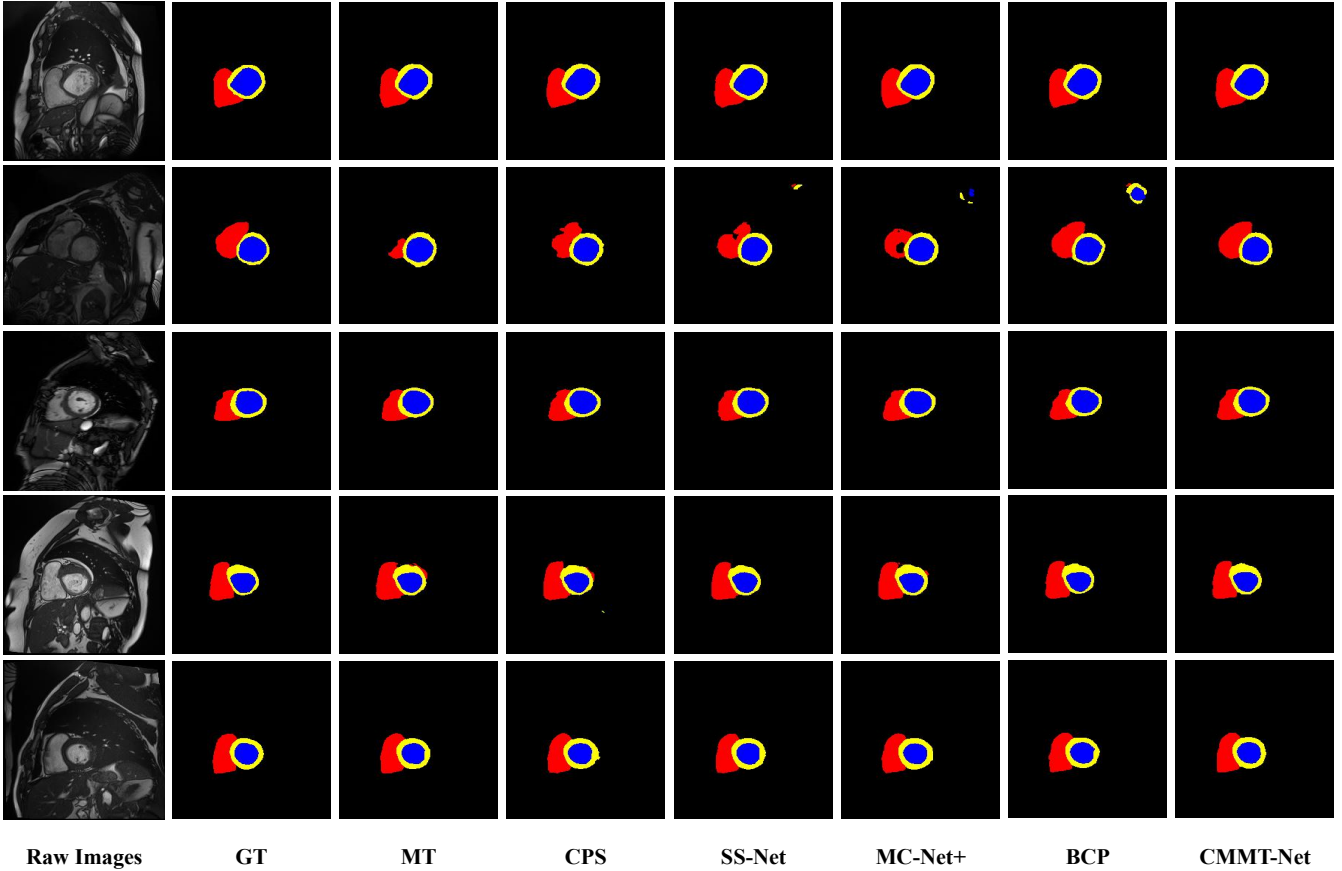}
	\caption{\textbf{Visualization results of different semi-supervised segmentation methods under 10\% labeled data for the ACDC dataset}.}
	\label{fig:acdc}
\end{figure*}

Figs.~\ref{fig:la},  ~\ref{fig:pct} and ~\ref{fig:acdc} display representative qualitative results obtained on the LA, Pancest-CT, and ACDC datasets, each with only 10\% labeled data for training. From  Figure~\ref{fig:la}, it is evident that our method excels at segmenting fine details of the target organ, particularly edges that are prone to misidentification (as seen in the first row) or omission (as in the second row). More specifically, compared to the SOTA method BCP~\cite{bai2023bidirectional}, our CMMT-Net achieves more accurate segmentation results (the first row), without overlooking smaller regions. BCP~\cite{bai2023bidirectional} produces misclassified regions and overlooks segmentation details (the second row). Similar observations can be made from  Figure~\ref{fig:pct}.  In the 2D segmentation task, MT yields subpar segmentation results, particularly struggling to segment the foreground, as demonstrated in the second row of ACDC images. Although SS-Net and MC-Net+ perform better than MT on ACDC, both methods misclassify the foreground (as seen in the second row). In contrast, our proposed CMMT-Net effectively rectifies many misclassified regions and captures previously overlooked segmentation details in the results obtained by other SSMIS methods, highlighting the effectiveness of CMMT-Net.

Additionally, we present the segmentation results of each component of our CMMT-Net in Figures~\ref{fig:la} and~\ref{fig:pct}. Notably, in Figure~\ref{fig:la} (the second row), the exclusion of CutMix results in the neglect of certain regions. When VAT and CCT are omitted, numerous regions are misclassified. Nevertheless, the combination of all these methods within our CMMT-Net yields markedly more accurate segmentation results.


\section{Discussions}
 
The utilization of cross-head co-training (CCT) and Mean-Teacher (MT) models, combined with consistency regularization and self-training, has found applications in various domains, including domain adaptation~\cite{yao2022enhancing} and learning with noisy labels~\cite{li2020dividemix}. Our CMMT-Net has the potential to extend the applicability of our approach to broader medical image domains, as exemplified by semi-supervised medical image classification. This extension provides an opportunity to leverage the inherent potential of limited labeled data for distinguishing between benign and malignant cases. This study holds the promise of benefiting numerous related fields that incorporate consistency regularization as part of their learning objectives. Additionally, we acknowledge that our approach may find direct applications in recent holistic schemes that combine pseudo-labeling and consistency regularization, such as the popular FixMatch~\cite{sohn2020fixmatch} framework designed for natural images. In FixMatch, one-hot pseudo labels from weakly perturbed unlabeled images are used to supervise the prediction of strongly perturbed unlabeled images with noticeable modifications. To harness the full potential of unlabeled data, another approach involves exploring synergies between our proposed Cross-head Mutual Mean-teaching Network and other techniques, such as Diffusion models~\cite{yang2022diffusion}, which could offer avenues for improved performance and robustness by generating more training samples. 

In terms of limitations,  to achieve feature-level perturbation, our current approach primarily adopts slightly different dual decoder, which  are pre-defined and limited to the availability of existing up-sampling strategies. As we expand our strategies, one natural extension is to apply Virtual Adversarial Training (VAT) at the feature level, potentially enhancing diversity between the dual decoders. Another avenue for exploration involves investigating disparities between different model architectures, such as Transformers and Convolutional Neural Networks (CNNs), to leverage the strengths of each architecture, although this may introduce additional complexity to the models. A crucial consideration is our reliance on model diversity in this strategy. One potential limitation arises if our various strategies do not effectively capture discriminative information from independent views. To address this limitation, we should explore different dimensions of co-training and systematically enhance diversity across various aspects. Additionally, gaining a theoretical understanding of how network homogenization impacts the generalization error in co-training is crucial. These areas offer promising opportunities for further research and development.

\section{Conclusion}
In this paper, instead of delving into the intricacies of complex network designs  for SSMIS studies, we propose a novel framework called CMMT-Net (Cross-head mutual mean-teaching Network). Our aim is to systematically enhance diversity at multiple levels, including data, features, and networks, ultimately improving SSMIS performance.  Specifically, CMMT-Net integrates both weak and strong augmentations within a cross-head co-training framework, combining the advantages of consistency and self-training. This approach not only enriches the diversity of samples used in consistency training but also addresses distribution bias stemming from differences between labeled and unlabeled data using Cross-set CutMix. Furthermore, it enhances self-training by improving the quality of pseudo-labels, reducing the influence of low-quality labels from peer models through two mean-teacher heads. Additionally, we introduce MVAT, which utilizes adversarial noise learned from the teacher model to perturb the inputs of the student model, thereby enhancing decision smoothing. Our detailed analysis
 highlights the crucial importance of perturbation and stabilization strategies in achieving impressive segmentation performance.  Experimental results on three publicly available datasets clearly demonstrate  that our approach outperforms previous SOTA methods by a significant margin across  various evaluation metrics.  Furthermore, our ablation studies confirm the contributions of each component of our approach.

\section{Acknowledgment}

This research was supported by the National Natural Science Foundation of China ( Grant No. 62376038).



\bibliographystyle{elsarticle-num} 
\bibliography{refs}






\end{document}